\documentclass[lettersize,journal]{IEEEtran}
\usepackage{amsmath,amsfonts}
\usepackage{algorithmic}
\usepackage{algorithm}
\usepackage{array}
\usepackage[caption=false,font=normalsize,labelfont=sf,textfont=sf]{subfig}
\usepackage{textcomp}
\usepackage{stfloats}
\usepackage{url}
\usepackage{verbatim}
\usepackage{graphicx}
\usepackage{cite}
\usepackage{indentfirst}

\begin{document}

\title{Past and Future Motion Guided Network for \\Audio Visual Event Localization}

\author{Tingxiu Chen, Jianqin Yin*, Jin Tang  
\thanks{This work was supported partly by the National Natural Science Foundation of China (Grant No. 62173045, 61673192), and partly by the Fundamental Research Funds for the Central Universities(Grant No. 2020XD-A04-2).}
}

\markboth{Journal of \LaTeX\ Class Files,~Vol.~14, No.~8, August~2021}%
{Shell \MakeLowercase{\textit{et al.}}: A Sample Article Using IEEEtran.cls for IEEE Journals}

\maketitle

\begin{abstract}
In recent years, audio-visual event localization has attracted much attention. It's purpose is to detect the segment containing audio-visual events and recognize the event category from untrimmed videos. Existing methods use audio-guided visual attention to lead the model pay attention to the spatial area of the ongoing event, devoting to the correlation between audio and visual information but ignoring the correlation between audio and spatial motion. We propose a past and future motion extraction (pf-ME) module to mine the visual motion from videos ,embedded into the past and future motion guided network (PFAGN), and motion guided audio attention (MGAA) module to achieve focusing on the information related to interesting events in audio modality through the past and future visual motion. We choose AVE as the experimental verification dataset and the experiments show that our method outperforms the state-of-the-arts in both supervised and weakly-supervised settings.\\
Available at: \url{http://tug.ctan.org/info/lshort/english/lshort.pdf}
\end{abstract}

\section{Introduction}
\IEEEPARstart{H}{uman} perceive the world through a variety of senses, such as hearing, vision, smelling, and touching.  Since the sound and vision are two dominant components, audio visual joint learning has attracted increasing attention in recent years. Audio visual event localization requires a machine to detect the event segment in an untrimmed video and recognize the category of the event. When an event occurs in both auditory and visual modes, it is regarded as an audio-visual event (as shown in Figure 1). Different from video action recognition\cite{ref20}, which directly recognizes the category from an untrimmed video, audio-visual event localization task\cite{ref1,ref2,ref3,ref4,ref5,ref6,ref7,ref8,ref9,ref10,ref11,ref12}  requires the computer to divide the video into some fixed length segments, recognize the category of each segment, and combine the adjacent segments with the same category as the detection result of localization. 

\begin{figure}
  \begin{center}
  \includegraphics[width=3.5in]{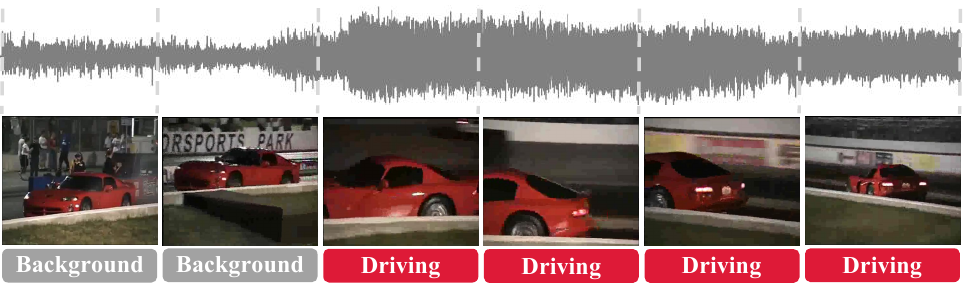}\\
  \caption{Illustration of audio visual event localization task. Audio visual events are defined as audible and visible events in the video(e.g.the first two video segment above is visible but Inaudible, so it’s a background segment). The purpose of audio-visual event localization is to locate the segments of audio visual events from the video and identify the categories of these events through mining the relationship between them while using the audio-visual modal information, which is very challenging but valuable.}\label{circuit_diagram}
  \end{center}
\end{figure}

\indent\setlength{\parindent}{2em}Audio visual event localization methods are required to consider the correlation between the two modalities. Existing methods rely on visual modality to provide spatial information and pick the audio motion as the dynamic information. They also use audio modality to guide the visual modality to pay attention to the space of interesting objects. However, there are some problems in these methods: (1)Existing audio visual event localization methods focus on the visual attention guided by hearing without mining the information related to interesting events in audio modality\cite{ref6,ref9,ref12}; (2) Existing methods extract dynamic information only from audio modality, which ignores the motion modeling of visual modality; We propose the past, future motion guided network (PFMGN) to pay attention to the information related to interesting events by screening the consistency information between dynamic vision and hearing.\\
\indent\setlength{\parindent}{2em}For selecting the relevant content of interested events in audio modality and reducing the influence of background noise, we propose motion-guided audio attention module. The motion in audio is rich, but the key motion content is not distilled, so the background noise interferes with the event localization. Visual spatial information is a static expression unable to resonate with the rich motion of voice. However, visual motion can be used as the guidance to retain the audio content highly related to the event to be detected, and reduce the negative impact of background noise to improve the detection performance. As the visual motion information is important for audio attention, we propose the motion-guided audio attention module, which exploits visual motion as the attention guidance of audio modality to focuse on auditory information highly consistent with visual motion and reduce the background noise.\\
\indent\setlength{\parindent}{2em}However, visual motion is implicitly expressed in the original visual modality. To express implicit motion information clearly , we propose the pf-ME (past and future motion excitation) module to exract the visual motion information. Visual motion information is usually represented by optical flow\cite{ref13} in some existing video understanding methods. However, the computation of optical flow is time-consuming and storage demanding. In recent years, lightweight motion extraction modules have been proposed to obtain motion information efficiently while slightly losing accuracy, such as ME\cite{ref14} and MS\cite{ref15}. However, they only focus on the future motion information and ignore the past one. To address the lack of the past motion in previous motion extraction module, we propose a new motion extraction module (pf-ME) to "see" the motion of the past, future, and current segments at the same time.\\
\indent\setlength{\parindent}{2em}To sum up, the contributions of this paper are as follows:\\
\indent\setlength{\parindent}{2em}(1) Visual motion information is introduced to mine the information related to audio event-related content by the motion-guided audio attention module we proposed. Meanwhile, the background noise is filtered.\\
\indent\setlength{\parindent}{2em}(2) A new motion extraction module: pf-ME is proposed to extract the past and future motion information to enrich visual motion content.\\
\indent\setlength{\parindent}{2em}(3) Experiments results demonstrate that our method outperforms the state-of-the-art in supervised and weakly-supervised tasks on the AVE dataset.\\
It's available at: \url{} which provides an overview of working with PFAGN.
\begin{figure*}
  \begin{center}
  \includegraphics[width=7.2in]{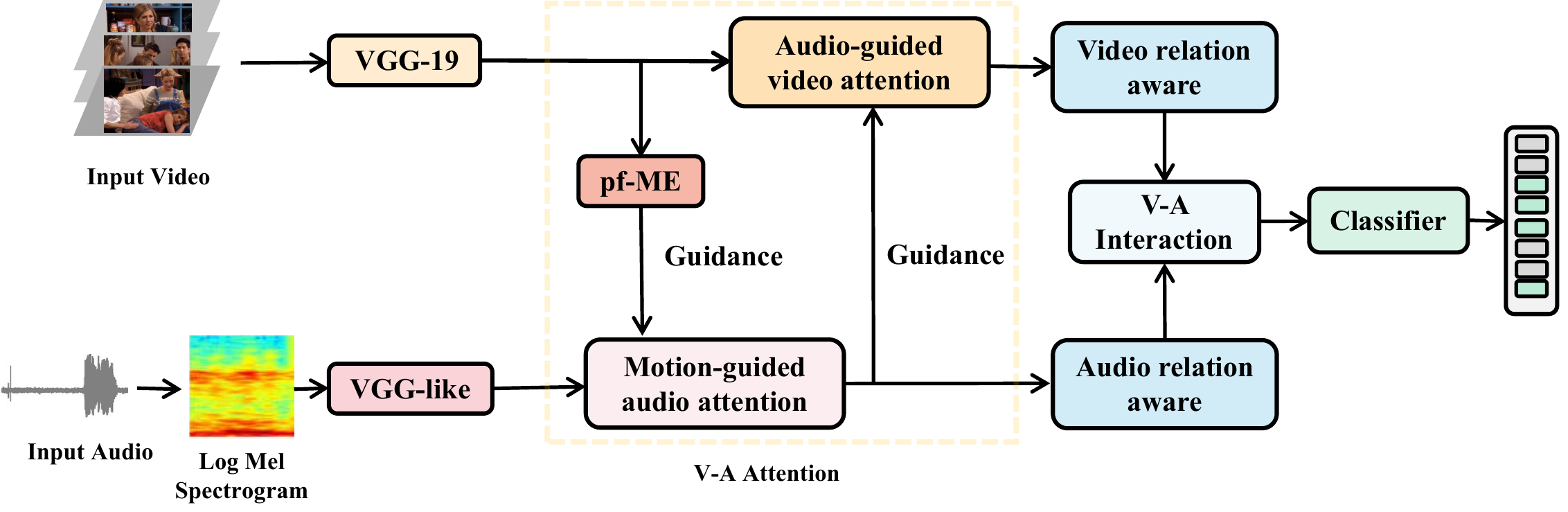}\\
  \caption{The pipeline of the proposed method for audio-visual event localization. The pf-ME module is proposed to extract past and future visual motion information as the guidance of audio attention. Then the motion-guided audion attention module is adopted to mine the event-related audio content while filtering background noise. Visual modality locate the spatial area around the vocal object benefit by the guidance of clean audio feature filtered by motion in the audio-guided video attention module. Finally, following the previous work, we obtain the relation-aware quality and integrate audio-visual features through the V-A interaction module for segment classification.}\label{circuit_diagram}
  \end{center}
\end{figure*}

\section{Related Works} 
\indent\setlength{\parindent}{2em}We first review the methods of existing audio-visual event localization, and then we narrow the scope to the audio guided visual attention audio visual event localization methods. Finally, two motion extraction modules in the action recognition task are discussed.\\
\indent\setlength{\parindent}{2em}{\bf{Audio visual event localization}}. The release of the AVE\cite{ref12} dataset promotes the research of audio-visual event localization tasks. Tian Y et al.\cite{ref12} proposed to learn event information through LSTM in a sequence to sequence manner. Hanyu X et al.\cite{ref6}proposed to learn inter and intra information between visual and audio modality by adaptive attention and self-attention modal. Jinxing Z et al.\cite{ref3} aggregates relevant information that probably not be available at the same time through the positive sample distribution model. And they all use the auditory guided visual attention module that we will discuss below.\\
\indent\setlength{\parindent}{2em}{\bf{Audio-guided visual attention}}. Some audio-visual joint learning methods use auditory guided visual attention to lead the model to follow the spatial area related to hearing. The sound source separation schemes proposed in \cite{ref18,ref19} show that the voices of different speakers can be distinguished by paying attention to the location of the spatial region around the speaker's voice and finding matching sound source information. In the audio-visual event localization task, \cite{ref12} firstly adopted the auditory guided visual attention mechanism, \cite{ref3,ref6,ref9} have followed this attention mechanism.\\
\indent\setlength{\parindent}{2em}The attention relies on visual modality to provide static spatial information and audio modality to provide dynamic information. However, the audio motion includes both event-related content and background noise. Treating event-related information indiscriminately with background noise will interfere with the judgment of event categories and create redundant calculations. We are committed to exploring ways to make audio modality focus on the event-related information to cope with the problem above. In addition to static spatial information, visual modality also contains motion content. Inspired by audio-guided visual spatial attention, turn around, the audio modality can also focus on event-related information through the visual motion-guided auditory attention mechanism, alleviating the interference of background noise.\\
\indent\setlength{\parindent}{2em}{\bf{Motion extraction modules}}. Visual motion information is not directly but implicit expressed in the video, so an effective motion information extraction module must be designed to explicitly model the visual motion information. Some audio-visual motion recognition methods have been committed to extracting visual motion through motion extraction modules in recent years. Compared with optical flow, they significantly reduce the calculation cost in the case of a small amount of accuracy wastage and provide favorable conditions for the practical application of the algorithm. \cite{ref15} proposed a motion squeeze (MS) extraction module to calculate the similarity between each pixel of the current frame and the pixel near the pixel of the next frame, using the max pool to retain the point with the highest similarity as the new position of the pixel, and fit the motion trajectory of the pixel. \cite{ref14} through the motion exception (ME) dynamic extraction module, the difference between two adjacent frames is used as the source of dynamic visual information. The CNN network is used to activate the dynamic, sensitive channel.\\
\indent\setlength{\parindent}{2em}Inspired by the ME module, we propose a pf-ME motion extraction module, which integrates the past and future motion of the current segment, activates the event-related motion in the past or future with CNN network and uses the visual motion extracted by this module as the guidance of audio attention.

\section{Method}
The framework of the method we propose is illustrated in Figure 2.Our algorithm is described as follows:\\
\indent\setlength{\parindent}{2em}1. First of all ,the architecture of our network is described, then the audio-visual feature extraction way. \\
\indent\setlength{\parindent}{2em}2. For visual motion extraction, we propose the pf-ME module to obtain the past and future visual motion as the guidance of audio attention, and the audio guided visual attention module (MGAA) for audio attention. The audio features distilled by visual motion are picked from the MGAA module.\\
\indent\setlength{\parindent}{2em}3. Following previous works, the audio features entry the audio-guided video attention module to follow the spatial area of vocal objects. Then the audio-visual relation-aware modules are used. The relation-aware feature is mined following the previous work\cite{ref9}, and we integrate audio-visual features through the V-A interaction module.\\
\indent\setlength{\parindent}{2em}4. Classifier and loss functions are introduced in turn. Overview of PFMG, based on CMRA \cite{ref9} network, we add the auditory attention branch guided by visual motion from the pf-ME module. The auditory guided spatial channel video attention module, relation-aware modules, and V-A interaction module are the same as in CMRA.\\
\indent\setlength{\parindent}{2em}{\bf{Audio and video features extraction}}.Firstly, an untrimmed video is divided into $T$ fixed-length segments without overlapping. For the audio modality, we convert the audio in every segment into spectrums through STFT (short-time Fourier transform)\cite{ref25} and map it to the Mel spectrum\cite{ref26} to leverage the model's perception of treble and bass frequency. For the visual modality, we adopt the first frame of each segment as the visual information of one segment to reduce redundant computation since the visual contents of adjacent frames are similar. Then the CNN networks are used to extract informative audio-visual features $ a \in \mathbb{R}^{ T\times d_{a}} $, $ v \in \mathbb{R}^{ T\times h\times w\times d_{v}} $.\\
\indent\setlength{\parindent}{2em}The audio-visual consistency information is extracted through the V-A attention module, where the attention of one modality is guided by another modality. In the motion-guided audio attention branch of visual motion guidance, the pf-ME module we proposed is to extract the dynamic visual information as the audio attention guidance. What's more, following the previous work, two relation aware modules are adopted to mine the intra and inter modalities' information through the cross-modality relation aware module. Then V-A interaction module integrates audio-visual features through the cross-modality relation aware module’s variant to integrate features of the two modalities and further explore the audio-visual interaction information. Finally, the characteristics of fusion are used to obtain the localization results.\\
\indent\setlength{\parindent}{2em}{\bf{Pf-ME visual motion extraction module}}. The visual feature extracted by the 2D-CNN of each segment represents the static visual information without mining dynamic visual change among these segments. Pf-ME module is proposed to extract visual motion features as audio modality's attention guidance to locate consistent audio-visual motion. The structure of the pf-ME module is described in detail, as shown in Figure 3.
Firstly, the visual features $ v:  v_{1},... v_{i},... v_{T} $ of $T$ segments enter the 2D-CNN to adjust the number of visual channels to be consistent with the audio channels. $ v' $ is a channel-adjusted result. The calculation formula is as follows: 

\begin{equation}
\label{deqn_ex1}
v' =  \text{conv}_{1} * {v}, \quad {v'}\in \mathbb{R} ^{T\times N\times d_{a}} 
\end{equation}

$ v_{i} \in \mathbb{R} ^{N\times d_{a}} $ is the first frame feature in the $i_{th}$ segment, $ N=h \times w $. Changes from $v_{i}$ through the 2D-CNN to $ v_{i-1}$ are regarded as the ‘past’ motion $M_{pi}$. Initial $M_{p1} = 0$. Similarly, we get $M_{fi}$ as ‘future’ motion, and last segment's motion : $M_{fT} = 0$. Pf-ME is inspired by ME\cite{ref14}, and 2D-CNNs are adopted to reduce the motion trajectory mismatch caused by pixel offset like ME. The calculation formula is as follows:

\begin{equation}
\label{deqn_ex1}
M_{pi} =  \text{conv}_{p} * {v_{i}} -{v_{i-1}}\\
\end{equation}

\begin{equation}
\label{deqn_ex1}
M_{fi} = \text{conv}_{f} * {v_{i+1}} -{v_{i}}
\end{equation}

\indent\setlength{\parindent}{2em}The "past" motion $M_{P}$ and the "future" motion $M_{F}$ are fused as $M_{S}$ to "see" motion changes of three segments. Then and avg-pooling in spatial dimension is adopted to obtain $M’$ to integrate the spatial information. Finally, the channel dimension of $M’$ is adjusted by 2D-CNN. The calculation formula is as follows:

\begin{align}
M_{s} &= M_{f}+M_{p},\quad M_{s}\in \mathbb{R}^{T\times h \times w \times d_{a}}
\end{align}

\begin{equation}
\label{deqn_ex1}
M' = \text{Avgpool}(M_{s}), \quad {M'}\in \mathbb{R}^{T\times d_{a}}
\end{equation}

\begin{equation}
\label{deqn_ex1}
M =  \text{conv}_{2} * {M'}, \quad {M}\in \mathbb{R} ^{T\times d_{a}} 
\end{equation}

\indent\setlength{\parindent}{2em}Pf-ME displays certain advantages in motion extracting. Firstly, compared with ME\cite{ref14}, MS\cite{ref15}, and other lightweight motion capture modules\cite{ref27,ref28}, the pf-ME module focuses on the dynamic changes of the first frame of adjacent segments instead of adjacent frames since adjacent segments have more obvious visual changes to reflect the event category. Secondly, the purpose of the pf-ME module is to simulate the "past" and "future" motion trajectories through a convolution kernel to activate the feature channels related to dynamic information in visual modal features. It simulates that we pay less attention to the static background than to the moving object when watching a video. The moving object is continuously determining the judgment of the event category in the video. the pf-ME module simulates the capture of moving objects as the human watching video by modeling the changes of motion trajectories in the "past" and "future."\\
\begin{figure}
  \begin{center}
  \includegraphics[width=2.5in]{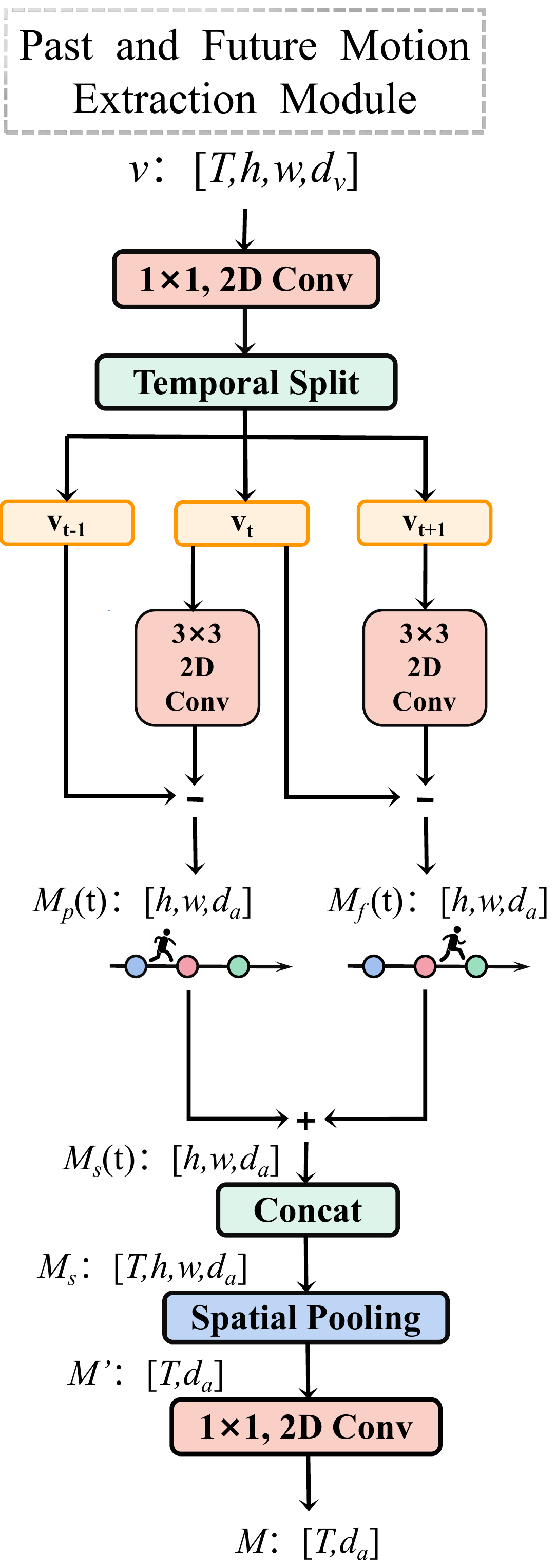}\\
  \caption{Illustration of the pf-ME module. pf-ME leverages CNNs to extract past and future visual motion as audio modality attention guidance.}\label{circuit_diagram}
  \end{center}
\end{figure}
\indent\setlength{\parindent}{2em}{\bf{V-A attention model}}. Audio-visual features are adopted as another modality's attention guidance in the V-A attention module to activate modality correlation information. Firstly, as shown in figure 4, we add a motion-guided audio attention (MGAA) branch with a temporal-wise attention module and a channel activation module to screen the information highly related to both audio and visual motion. After obtaining a new auditory feature $a'$ from the motion-guided audio attention module, the module attracts more attention to audio data related to the events category while reducing the impact of background noise. The calculation formulas in temporal attention module are as follows.\\
\begin{equation}
\label{deqn_ex1}
M_{ta} = \text{softmax}(\text{W}_{ta}M)  ,\quad M_{ta}\in \mathbb{R}^{T\times 1}\\
\end{equation} 
 
\begin{equation}
\label{deqn_ex1}
a_{1} =  a \oplus (M_{ta}\odot {a}) 
\end{equation} 

And the calculation formulas in channel attention module are as follows:\\ 
\begin{equation}
\label{deqn_ex1}
M_{ca} = \text{sigmoid}(M)  ,\quad M_{ca}\in \mathbb{R}^{T \times d_{a}}
\end{equation} 

\begin{equation}
\label{deqn_ex1}
a' =  a_{1} \oplus (M_{ca}\odot a_{1}) 
\end{equation} 
\indent\setlength{\parindent}{2em}In the existing methods, audio-guided visual attention is used to locate the spatial area of audible objects, only modelling the relation between audio and static vision. However, audio contains rich dynamic information, and the relationship between audio and dynamic vision needs to be fully explored. Temporal attention module in motion-guided audio attention module thar we proposed can distinguish whether a segment contains an action or not: segments with strong motion in both dynamic vision and audio are segments in which actions occur, while segments with weak visual motion are most likely to be segments presenting background noise even though the audio motion is strong. In the channel attention module, audio and dynamic visual resonance information is enhanced to reduce the influence of background noise and facilitate the classification of action categories.\\
\indent\setlength{\parindent}{2em}Then $a'$ is regarded as static visual guidance to pay attention to the event-related spatial area. Continuing the practice in the CMRAN , the original static visual feature v enters the channel attention module to obtain the static visual $v_{c'}$ that activates the event-related channel. The calculation formula is as follows:\\

\begin{equation}
\label{deqn_ex1}
v_{c} =  \text{ReLU}(\text{W}_{c1}{a'}) \odot \text{ReLU}(\text{W}_{c2}{v})  
\end{equation} 

\begin{equation}
\label{deqn_ex1}
v'_{c} =  \text{avg}({v_{c}})  \odot \text{W}_{c3}{v}
\end{equation}

\indent\setlength{\parindent}{2em}Then the spatial attention module obtains the static visual feature $v_{cs}'$ that focuses on the spatial features of area related to audio modality. The calculation formula is as follows:\\

\begin{equation}
\label{deqn_ex1}
v_{cs} =  \text{ReLU}(\text{W}_{s1}{a'}) \odot \text{ReLU}(\text{W}_{s2}v'_{c}) 
\end{equation} 

\begin{equation}
\label{deqn_ex1}
v'_{cs} =  \text{tanh}({v_{cs}})\text{W}_{s3} \otimes {v'_{c}} ,\quad{v'_{cs}}\in \mathbb{R}^{T \times d_{v}}
\end{equation} 

\indent\setlength{\parindent}{2em}Due to the similar spatial characteristics of adjacent frames, we select the first frame image of video segments as static visual information to represent spatial content against the visual motion. The audio spectrogram is adopted as the content of the audio modality carrying rich audio motion without static audio since there is no spatial structure in audio. However, the audio modality only interacts with static visual information, ignoring previous methods' visual motion modeling and interaction between audio and visual motion. In contrast, the audio guided by visual motion can retain the info highly related to audio-visual events. As the guidance of static vision, it is the joint guidance of dynamic sight and dynamic hearing. We model the visual motion information to interact with the audio and guide the network to pay attention to the information related to the event category through the V-A attention module.\\

\begin{figure*}
  \begin{center}
  \includegraphics[width=7in]{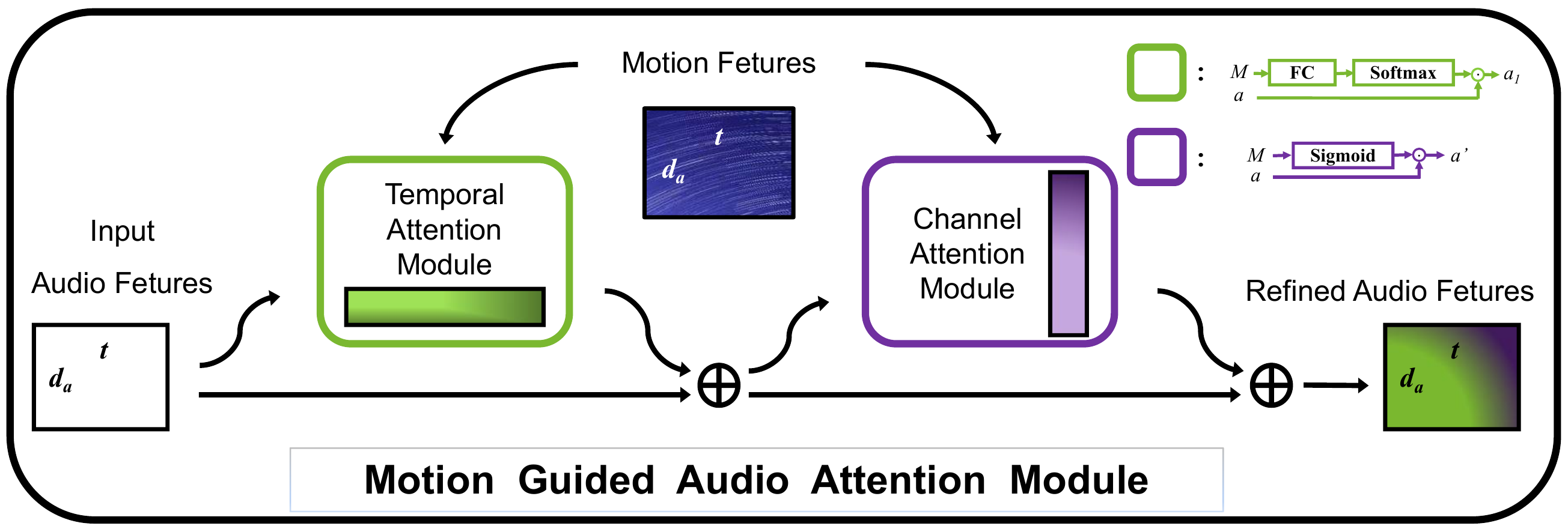}\\
  \caption{Illustration of the proposed motion-guided audio attention module. The green bar represents the temporal-wise attention module, the purple bar represents the channel-wise attention module, and the corresponding module details are in the upper right corner. Under the guidance of visual motion, the event-related information of audio modality is concerned, and the background noise is filtered out.}\label{circuit_diagram}
  \end{center}
\end{figure*}

\indent\setlength{\parindent}{2em}{\bf{RA(relation aware) module}}. After obtaining $a'$and $v_{cs}'$, continue the practice of CMRAN, we input audio-visual features into two-way RA (relationship aware) modules, respectively. The RA(relation aware) modules learn the correlation information within and with the help of the multimodal attention module between modalities. The calculation formula of the CMRA multimodal attention module is as follows:

\begin{equation}
 \text{CMRA}(x,y) = \text{softmax}(\frac {q_{1}(K_{1,2})^{T}}{d_{m}})X_{1,2}
\end{equation}

\begin{equation}
\label{deqn_ex1}
q_{1} = x\text{W}^{Q},K_{1,2} = g_{x,y}\text{W}^{K},X_{1,2} = g_{x,y}\text{W}^{X}
\end{equation} 

\begin{equation}
\label{deqn_ex1}
g_{xy} = \text{Concat}(x,y)
\end{equation} 

The calculation formula of visual mode branch RA module is as follows:

\begin{equation}
V_{R} = \text{CMRA}(V_{cs},A')
\end{equation}

The calculation formula of audio relation aware is as follows:

\begin{equation}
A_{R} = \text{CMRA}(A',V_{cs})
\end{equation}

\indent\setlength{\parindent}{2em}Then, the audio-visual interaction is further carried out through the relation aware module. The audio-visual feature splicing result $O_{av}$ and point multiplication result $f_{av}$ are used as new input to the CMRA model. The resonance channel between vision and acoustics is activated by combining audio-visual features to obtain feature O. Finally, Of is obtained from $O$ and $Oav$ as the classification feature.
\begin{equation}
\label{deqn_ex1}
F_{av} = A_{R}\odot V_{R}
\end{equation} 

\begin{equation}
\label{deqn_ex1}
O_{av} =\text{Concat}(A_{R} , V_{R})
\end{equation} 

\begin{equation}
\label{deqn_ex1}
O =\text{CMRA}(F_{av}, O_{av})
\end{equation} 

\begin{align}
O_{f} &= O \oplus O_{av}
\end{align} 

\indent\setlength{\parindent}{2em}{\bf{Classifier and loss}}. $O_{av} \in \mathbb{R}^{T \times D_{v}}$ is the final feature for classification. In the supervised method, the segment level labels are provided, and in the weakly-supervised method, only the category of the whole video is provided. FC is used as a classifier to locate the event segments and predict the event categories.
The audio-visual correlation score Se represents the audio-visual correlation degree of each segment, and Sc represents the video level category obtained through the classifier.
As a complete video classification feature, $ O_{av} \in \mathbb{R}^{1\times D_{v}} $is obtained by Max-pooling from $O_{av}$ in the time dimension. The calculation formula is as follows:

\begin{equation}
\label{deqn_ex1}
O_{av} = \text{Maxpool}(O_{av}) \quad,{O_{av}}\in \mathbb{R}^{1\times d_{v}} 
\end{equation}

\indent\setlength{\parindent}{2em}$S_{e}$ represents the audio-visual correlation degree score of a video segment. When the $S_{e}$ of a video segment is greater than 0.5, the segment is an audio-visual event occurrence segment. Since most videos contain only one event category, the category of the complete video is considered as the category for all audio-visual related segments. When the score $S_{e}$ is less than 0.5, the fragment is considered as a background fragment. $S_{c}$ represents the category results for the complete video. Calculated as follows:
\begin{equation}
\label{deqn_ex1}
S_{c} = \text{Softmax}(O_{av}W_{c})
\end{equation} 

\begin{equation}
\label{deqn_ex1}
S_{e} = \text{sigmoid}(O_{av}W_{e})
\end{equation}  

\indent\setlength{\parindent}{2em}The complete video category detection results are constrained by multi-classification CE (cross-entropy) loss, and the audio-visual correlation detection results of fixed length segments are constrained by binary classification CE loss. The final loss is the sum of the two. The loss is calculated as follows: 

\begin{equation}
\label{deqn_ex1}
L_{c} = -\sum_{k=1}^{C} y_{i}\text{log}(S_{ci})
\end{equation}

\begin{equation}
\label{deqn_ex1}
L_{e} = -y\text{log}(S_{e})-(1-y)\text{log}(1-S_{e})
\end{equation}

\begin{equation}
\label{deqn_ex1}
L = L_{c} +L_{e}
\end{equation}

\indent\setlength{\parindent}{2em}Where $y_{i}$ is the label and $p_{i}$ is the prediction result.In the supervised task, segment level labels are provided in training, including audio-visual correlation labels of each segment and category labels of each segment. In weakly-supervised training, only the category of the complete video is provided, and the supervision signals are mined from the data (such as whether audio and video exist at the same time). We adapt the label by adding the prediction results of each segment in time dimension element-wise.

\section{Experiment} 
\indent\setlength{\parindent}{2em}{\bf{Dataset}}. We perform extensive experiments on the AVE dataset. The AVE dataset\cite{ref12} contains 4143 videos on YouTube, 66.4$ \text{\% } $of which are 10s long, and the rest are no less than 2s long, covering 28 event categories. It involves various audio-visual events such as human activities, animal activities, music performance, and vehicle sound. It provides a dataset with a wide range and samples full of time inconsistency and mutation. The data scene is rich and close to the actual event scene. AVE is the most commonly used public dataset for audio-visual event localization. Consistent with the practice of \cite{ref1,ref2,ref3,ref4,ref5,ref6,ref7,ref8,ref9,ref10,ref11,ref12}, we use accuracy to measure the performance of event localization.\\
\indent\setlength{\parindent}{2em}{\bf{Implementation Details}}. We adopt 2D-CNN networks that pre-trained on large datasets to extract audio and visual features of video segments. We adopt VGG-19 pre-trained on ImageNet\cite{ref24} to extract the visual feature. The extracted video features are $v \in \mathbb{R} ^ {T \times h \times w \times d_{v}}$, $T$ is the number of fixed- length video segments segmented from untrimmed videos. A video is divided into some fixed-length segments and takes the first frame of each segment as visual information. Therefore, one video obtains $T$ frames, and the extracted feature is $T$ length, $h$ and $w$ are the length and width of each segment's first frame feature map, and $d_{v}$ is the visual channel dimension. As far as the audio feature, we use VGG-like feature extraction network pre-trained on AudioSet dataset. The extracted feature $a \in \mathbb{R} ^{T \times d_{a}}$ is the spectrogram feature of audio in every segment, and $d_{a}$ is the audio feature channel dimension. The verification experiment is based on PyTorch, \rm{batch\_size} = 32, and learning rate $l$ = $5 \times 10^{-4}$.\\
\begin{table}
\begin{center}
\caption{Comparisons with state-of-the-arts in supervised task on AVE dataset (* indicates the reproduced performance)\label{tab:table1}}
\label{tab1}
\begin{tabular}{| c | c | c | c | c |}
\hline
Method & Audio Feature & Video Feature & Accuracy(\text{\% })\\
\hline
AVEL\cite{ref12} & VGG-19 & VGG-like & 72.7 \\
\hline
AVIN\cite{ref11} & VGG-19 & VGG-like & 75.2 \\ 
\hline
AVSDN\cite{ref10} & VGG-19 & VGG-like & 75.4 \\
\hline 
CMRAN*\cite{ref9} & VGG-19 & VGG-like & 76.8 \\
\hline 
CMRAN\cite{ref9} & VGG-19 & VGG-like & 77.4 \\
\hline 
PSP\cite{ref3} & VGG-19 & VGG-like & 77.8 \\
\hline 
Ours & VGG-19 & VGG-like & {\bf{78.5}} \\
\hline 
\end{tabular}
\end{center}
\end{table}
\begin{table}
\begin{center}
\caption{Comparisons with state-of-the-arts in weakly-supervised task on AVE dataset (* indicates the reproduced performance)\label{tab:table1}}
\label{tab1}
\begin{tabular}{| c | c | c | c | c |}
\hline
Method & Audio Feature & Video Feature & Accuracy(\text{\% })\\
\hline
AVEL\cite{ref12} & VGG-19 & VGG-like & 66.7 \\
\hline
AVIN\cite{ref11} & VGG-19 & VGG-like & 69.4 \\ 
\hline
AVSDN\cite{ref10} & VGG-19 & VGG-like & 74.2 \\
\hline 
CMRAN*\cite{ref9} & VGG-19 & VGG-like & 72.5 \\
\hline 
CMRAN\cite{ref9} & VGG-19 & VGG-like & 72.9 \\
\hline 
PSP\cite{ref3} & VGG-19 & VGG-like & {\bf{73.5}} \\
\hline 
Ours & VGG-19 & VGG-like & {\bf{73.5}} \\
\hline 
\end{tabular}
\end{center}
\end{table}

\begin{figure*}
  \begin{center}
  \includegraphics[width=7.2in]{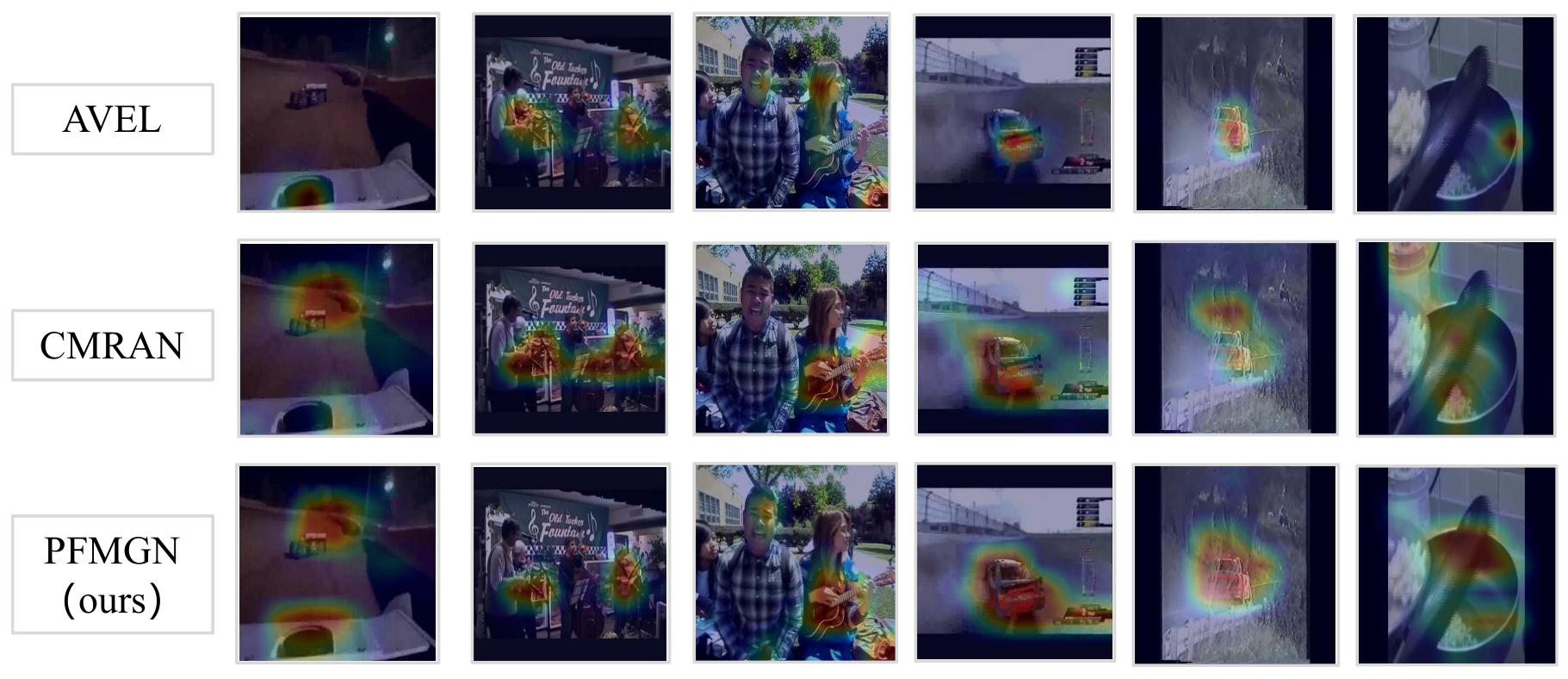}\\
  \caption{Qualitative examples of different attention methods. The top row, middle row and bottom row show the visualization examples of CMRAN and our method respectively. As can be seen from the illustration, our PFMG network is more focused than AVEL in locating audible objects and more accurate than CMRAN. This is because hearing, after being screened by visual motion, retains information highly relevant to audio-visual events, thus better directing the visual attention area.}\label{circuit_diagram}
  \end{center}
\end{figure*}

\indent\setlength{\parindent}{2em}{\bf{Comparisons with state-of-the-arts}}. We compare our PFAG network with state-of-the-art methods on the AVE dataset in supervised and weakly-supervised AVE localization(as shown in Table 1. For a fair comparison, we compare the localization results of the PFAG network with previous methods with the same audio-visual features. For a fair comparison, we relive CMRAN on the same platform, we get 1.7\text{\% } accuracy higher in the supervised settings, and 1\text{\% } higher in the weakly-supervised settings, which proves the effectiveness of introducing visual motion information and the visual motion extraction ability pf-ME module.\\
\indent\setlength{\parindent}{2em}In present methods, the audio guided spatial region attention mechanism has a good performance in the audio-visual event localization task. It’s benefit from the focus on the event-related spatial regions and the reduction of attention to the background spatial information under the guidance of hearing. However, they ignore the distillation of audio information. Auditory signals have rich motion information, but contain background noise that interferes with event localization. The motion information in visual modality is able to be the basis for the screening of audio modality. it is that most of the features with strong audio-visual dynamic are event-related features. If the audio features are dynamic but the corresponding visual features are relatively static, these audio features are most likely to point to the background noise content with interference. And the visual and audio features of rich dynamic information represent the key information related to the event.\\
\indent\setlength{\parindent}{2em}We propose the PF-ME module to mine the visual motion from original videos that implicitly expresses the motion, so that it can be expressed explicitly. Experiments show the effectiveness of PF-ME module in motion extraction. Then, the extracted visual motion is used as auditory guidance to filter key information related to the event in audio modality and filter the background noise, so as to improve the event localization performance of the proposed method. Experiments demonstrate the positive effects of introducing visual motion.\\
\indent\setlength{\parindent}{2em}{\bf{Ablation Study}}. We split the proposed framework to verify the validity of the different modules. Firstly, We discuss the utility of visual motion, then the function of past visual motion information in the pf-ME module. VGG-19 and VGG-like are also used as audio-visual feature extractors, the same as existing methods.\\
The utility of visual motion. In order to prove the effectiveness of visual motion to the network, we compare RNN and ME modules to extract dynamic visual information, which is adopted as the attention guidance information of audio to pay attention to content related to the event (as shown in figure 2).\\
\indent\setlength{\parindent}{2em}Compared with the visual modality, the auditory mode does not have a spatial structure. Still, it contains rich, dynamic information, which can make up for the lack of dynamic modeling of visual mode and complement the rich spatial information.\\
\begin{table}
\begin{center}
\caption{Ablation study on supervised task on AVE dataset (* indicates the reproduced performance)\label{tab:table1}}
\label{tab1}
\begin{tabular}{| c | c | }
\hline
Method & {Accuracy(\text{\% })}\\
\hline
CMRAN*\cite{ref9}& 76.841  \\
\hline 
+RNN(w/o temporal attention)& 76.692  \\
\hline 
+motion from the ME (w/o temporal attention)& 77.313 \\
\hline 
+motion from the pf-ME (w/o temporal attention)& 77.488  \\
\hline 
+motion from the pf-ME (w temporal attention)& {\bf{78.458}} \\
\hline 
\end{tabular}
\end{center}
\end{table}
\begin{table}
\begin{center}
\caption{Ablation study on weakly-supervised task on AVE dataset (* indicates the reproduced performance)\label{tab:table1}}
\label{tab1}
\begin{tabular}{| c | c |}
\hline
Method &{Accuracy(\text{\% })}\\
\hline
CMRAN*\cite{ref9}& 72.488 \\
\hline 
+RNN(w/o temporal attention)& 72.355 \\
\hline 
+motion from the ME (w/o temporal attention)&  72.910 \\
\hline 
+motion from the pf-ME (w/o temporal attention)& {\bf{73.532}} \\
\hline 
+motion from the pf-ME (w temporal attention)&  73.159 \\
\hline 
\end{tabular}
\end{center}
\end{table}
\indent\setlength{\parindent}{2em}However, part of the rich motion of the audio modality is background noise (such as background music, street noise, wind, and rain, etc.), but the audio motion highly related to visual information is the critical information of audio-visual event localization. We are committed to retaining the auditory information related to events and filtering the background noise in the audio modality. The feature of high correlation between audio and visual motion is highly associated with audio-visual events, respectively. Experiments show that visual motion extracted by the ME module to guide the auditory modal attention model improves the accuracy of supervised localization by 0.48$ \text{\% } $ and weakly-supervised localization by 0.42$ \text{\% } $, which verifies the effectiveness of introducing the visual motion.\\
\indent\setlength{\parindent}{2em}The screening of visual motion information guides audio pays attention to the characteristics highly related to the event and suppresses the background sound characteristics irrelevant to the event (such as street noise, wind, and rain, background music). Then input the audio feature $a'$ purified by dynamic visual attention to guide the concentration of static vision $v$ to obtain a new visual feature $v'$, which drives the network to pay more attention to the spatial area of the vocal source. The purification of audio features avoids some attention to the spatial features of background noise to better focus on the spatial content related to the concerned events.\\
\indent\setlength{\parindent}{2em}The utility of extracting "past" visual motion. Based on verifying the effectiveness of visual motion, we make an in-depth study on how to extract effective visual motion. Results of the ME module extracting visual motion (see Table 2 "+ motion (me)" results) show that the addition of visual motion is beneficial to the localization of audio-visual events. However, the ME module only captures the future visual motion information.\\
\indent\setlength{\parindent}{2em}Simulate the process of human eyes watching a video to obtain information. People often understand the "antecedents" and "consequences" of events through a dynamic video in the past and the future and judge the events in the current video clip. Therefore, this paper proposes the pf-ME module with past and future visual motion to capture the complete event process and reduce ambiguity through the current video segment and the visual-spatial changes in the previous and subsequent periods. Results show that compared with the ME module, which only extracts future dynamic information, the pf-ME module obtains past and future dynamic information at the same time, which improves the accuracy by 0.18\text{\% } in supervised localization and 0.2\text{\% } in weakly-supervised localization, which proves the effectiveness of extracting the past information.\\
\section{Conclusion }
\indent\setlength{\parindent}{2em}In this paper, we explore the efficient audio-visual event localization method through motion guided audio attention mechanism. Based on the auditory information guidance network paying attention to the spatial area of vocals, the visual motion is introduced to train the network to pay attention to the highly relevant features of events in the audio modality. A pf-ME module for extracting visual motion is proposed to make the network "see" the past and future motion as the guidance of audio modality attention. Conclusion and persuasive experimental results show that our method achieves SOTA accuracy, which proves the effectiveness of our approach.
In the future, audio-visual joint learning methods can adapt multi-channel audio information collected by multi microphones and combine the position information of sound receiving microphones as spatial information of audio modality. And visual-spatial information to more accurately locate the vocal object and improve the accuracy of event localization.\\

\newpage

\vspace{11pt}

\begin{IEEEbiography}[{\includegraphics[width=0.8in,height=1in,clip,keepaspectratio]{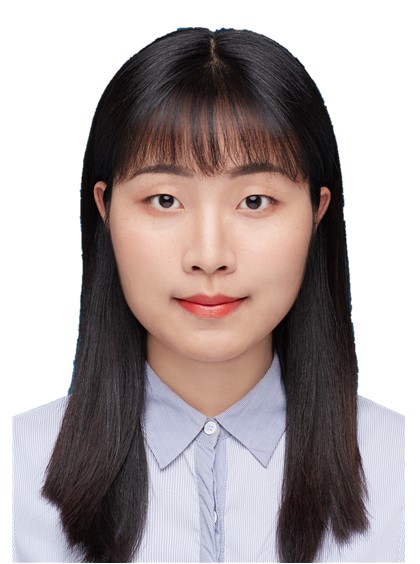}}]{Tingxiu Chen}
Master at Beijing University of Posts and Telecommunications. Her research interests include computer vision, deep learning and machine learning.Email: ctx2333@bupt.edu.cn
\end{IEEEbiography}

\vspace{11pt}

\begin{IEEEbiography}[{\includegraphics[width=0.8in,height=1in,clip,keepaspectratio]{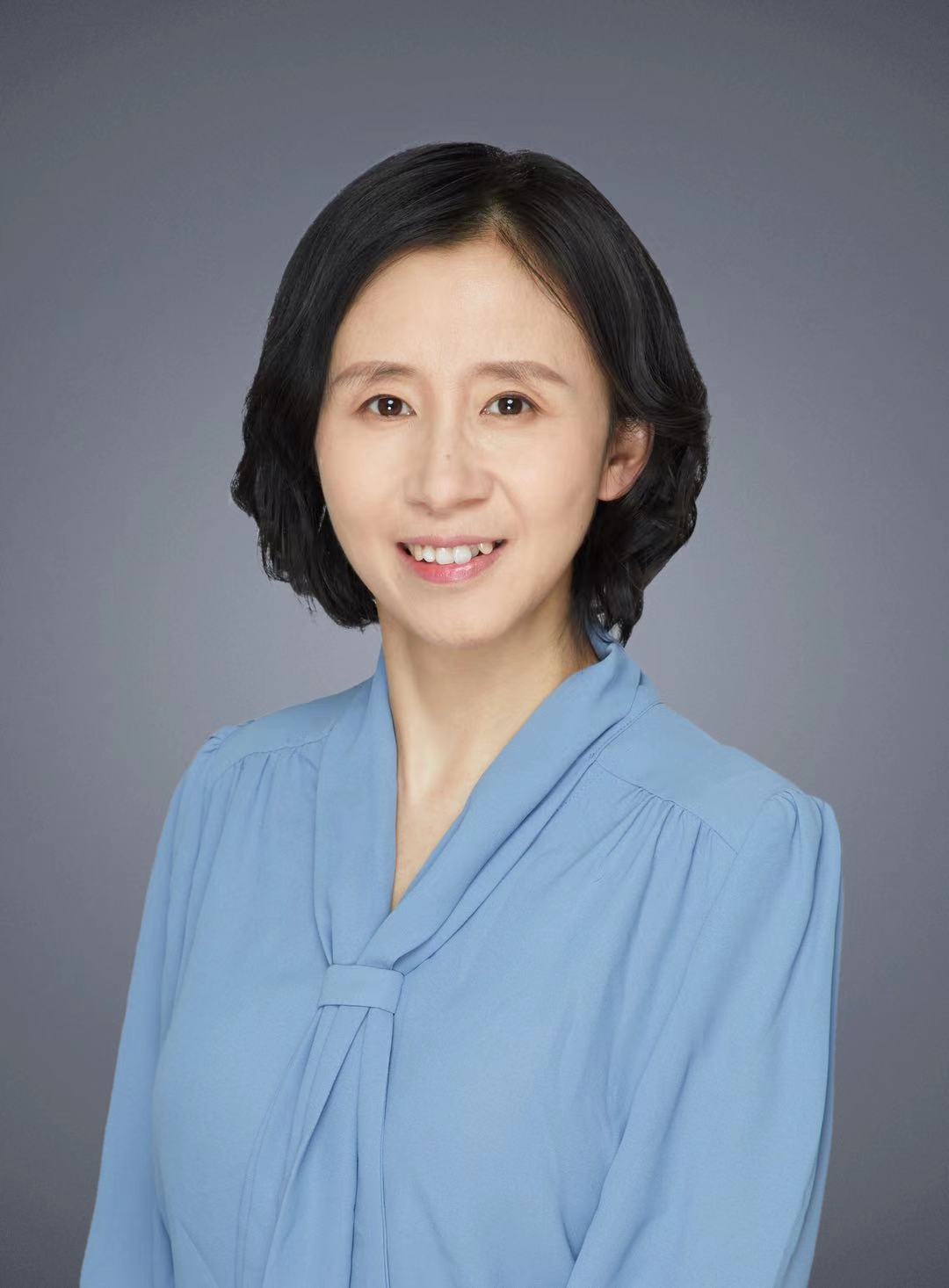}}]{Jianqin Yin}
received the Ph.D. degree from Shandong University, Jinan, China, in 2013. She currently is a Professor with Artificial Intelligence School, Beijing University of Posts and Telecommunications, Beijing, China. Her research interests include service robot, pattern recognition, machine learning and image processing. Email: jqyin@bupt.edu.cn.
\end{IEEEbiography}

\begin{IEEEbiography}[{\includegraphics[width=0.8in,height=1in,clip,keepaspectratio]{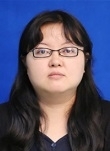}}]{Jin Tang}
 received the Ph.D. degree from  Beijing Institute of Technology, Beijing, China, in 2007. currently is a Assistant Professor with Artificial Intelligence School, Beijing University of Posts and Telecommunications, Beijing, China.  Her research interests include signal processing, pattern recognition, and deep learning. Email: tangjin@bupt.edu.cn

\end{IEEEbiography}

\vfill


\begin{thebibliography}{1}
\bibliographystyle{IEEEtran}
\bibitem{ref1}
H. Xuan, L. Luo, Z. Zhang, J. Yang and Y. Yan, ``DMMAN: a two-stage audio-visual fusion framework for sound separation and event localization,'' in \textit{Neural Networks}, 2021, vol.133, pp.  229--239.

\bibitem{ref2}
H. Xuan, L. Luo, Z. Zhang, J. Yang and Y. Yan, ``Discriminative cross-modality attention network for temporal inconsistent audio-visual event localization,'' in \textit{IEEE Transactions on Image Process}, vol.30, pp. 7878--7888, 2021.

\bibitem{ref3}
J. Zhou, L. Zheng, Y. Zhong, S. Hao and M. Wang, ``Positive sample propagation along the audio-visual event line,'' in \textit{Proceedings of the IEEE conference on computer vision and pattern recognition}, 2021, pp. 8436--8444.

\bibitem{ref4}
J. B. Li, K. Ma, S. Qu, P. Huang, F. Metze, ``Audio-visual event recognition through the lens of adversary,'' in \textit{Proceedings of the IEEE conference on international conference on acoustics, speech and signal processing}, 2021, pp. 616--620.

\bibitem{ref5}
B. Duan, H. Tang, W. Wang, Z. Zong, G. Yang, Y. Yan, ``Audio-visual event localization via recursive fusion by joint co-attention,'' in \textit{Proceedings of the IEEE workshop on applications of computer vision}, 2021, pp. 4012--4021.

\bibitem{ref6}
H. Xuan, Z. Zhang, S. Chen, J. Yang and Y. Yan, ``Cross-modal attention network for temporal inconsistent audio-visual event localization,'' in \textit{Proceedings of the AAAI conference on artificial intelligence}, 2020, pp. 279--286.

\bibitem{ref7}
Yan-Bo Lin, Yu-Chiang Frank Wang, ``Audiovisual transformer with instance attention for audio-visual event localization,'' in \textit{Proceedings of the asian conference on computer vision}, Springer, 2020, pp. 274--290.

\bibitem{ref8}
J. Ramaswamy, ``What Makes the Sound?: A dual-modality interacting network for audio-visual event localization,'' in \textit{Proceedings of the IEEE conference on international conference on acoustics, speech and signal processing}, 2020, pp. 4372--4376.

\bibitem{ref9}
H. Xu, R. Zeng, Q. Wu, M. Tan, C. Gan, ``Cross-modal relation-aware networks for audio-visual event localization,'' in \textit{Proceedings of the ACM Multimedia}, 2020, pp. 3893--3901.

\bibitem{ref10}
Yan-Bo Lin, Yu-Jhe Li, Yu-Chiang Frank Wang, ``Dual-modality seq2seq network for audio-visual event localization,'' in \textit{Proceedings of the IEEE conference on international conference on acoustics, speech and signal processing}, 2019, pp. 2002--2006.

\bibitem{ref11}
Y. Wu, L. Zhu, Y. Yan and Y. Yang, ``Dual attention matching for audio-visual event localization,'' in \textit{Proceedings of the IEEE conference on international conference on computer vision}, 2019, pp. 6291--6299.

\bibitem{ref12}
Y. Tian, J. Shi, B. Li, Z. Duan and C. Xu, ``Audio-visual event localization in unconstrained videos,'' in \textit{Proceedings of the europeon conference on computer vision}, Springer, 2018, pp. 252--268.

\bibitem{ref13}
K. Simonyan and A. Zisserman, ``Two-stream convolutional networks for action recognition in videos,'' in \textit{in Proceedings of the 27th International Conference on Neural Information Processing Systems}, 2014, pp. 568--576.

\bibitem{ref14}
Y. Lin, Y. Li and Y. F. Wang, ``TEA: temporal excitation and aggregation for action recognition,'' in \textit{Proceedings of the IEEE conference on computer vision and pattern recognition}, 2020, pp. 906--915.

\bibitem{ref15}
H. Kwon, M. Kim, S. Kwak, M. Cho, ``MotionSqueeze: neural motion feature learning for video understanding,'' in \textit{Proceedings of the europeon conference on computer vision}, Springer, 2020, pp. 345--362.

\bibitem{ref16}
Y. Cheng, R. Wang, Z. Pan, R. Feng and Y. Zhang, ``Look, listen, and attend: co-attention network for self-supervised audio-visual representation learning,'' in \textit{Proceedings of the ACM Multimedia}, 2020, pp. 3884--3892.

\bibitem{ref17}
A. Ephrat, I. Mosseri, O. Lang, T. Dekel, K. Wilson, A. Hassidim, W. T. Freeman and M. Rubinstein, ``Looking to listen at the cocktail party: a speaker-independent audio-visual model for speech separation,'' in \textit{ACM Trans. Graph}, 2019, vol.37, no.4, pp. 112:1--112:11.

\bibitem{ref18}
C. Gan, D. Huang, H. Zhao, J. B. Tenenbaum and A. Torralba, ``Music gesture for visual sound separation,'' in \textit{Proceedings of the IEEE conference on computer vision and pattern recognition}, 2020, pp. 10475--10484.

\bibitem{ref19}
P. Pareek and A. Thakkar, ``A survey on video-based human action recognition: recent updates, datasets, challenges, and applications,'' in \textit{Artif. Intell. Rev}, 2021, vol.54, no.3, pp. 2259--2322.

\bibitem{ref20}
S. Adavanne, P. Pertilä, T. Virtanen, ``Sound event detection using spatial features and convolutional recurrent neural network,'' in \textit{Proceedings of the IEEE conference on international conference on acoustics, speech and signal processing}, 2017, pp. 771--775.

\bibitem{ref21}
Y. Cao, T. Iqbal, Q. Kong, F. An, W. Wang, M. D. Plumbley, ``An improved event-independent network for polyphonic sound event localization and detection,'' in \textit{Proceedings of the IEEE conference on international conference on acoustics, speech and signal processing}, in , 2021, pp. 885--889.

\bibitem{ref22}
T. Afouras, A. Owens, J. S. Chung and A. Zisserman, ``Self-supervised learning of audio-visual objects from video,'' in \textit{Proceedings of the europeon conference on computer vision}, Springer, 2020, pp. 208--224.

\bibitem{ref23}
Gao R, Oh T, Grauman K, Torresani L, ``Listen to look: action recognition by previewing audio,'' in \textit{Proceedings of the IEEE conference on computer vision and pattern recognition}, 2020, pp. 10454-10464.

\bibitem{ref24}
Jia D, Wei D, Richard S, Li-Jia L, Kai L, Li F, ``ImageNet: A large-scale hierarchical image database,'' in \textit{Proceedings of the IEEE conference on computer vision and pattern recognition}, 2009, pp. 248-255.

\bibitem{ref25}
Kootsookos P.J, Lovell B.C, Boashash B, ``A unified approach to the STFT, TFDs, and instantaneous frequency,''in \textit{IEEE Transactions on Signal Processing} , 2002, vol.40, no.8, pp. 1971-1982.

\bibitem{ref26}
S. Molau, M. Pitz, R. Schlüter, H. Ney,``Computing Mel-frequency cepstral coefficients on the power spectru'','in \textit{Proceedings of the IEEE conference on international conference on acoustics, speech and signal processing} 2001: 73-76

\bibitem{ref27}
Y. Zhang, L. Niu, Z. Pan, M. Luo, J. Zhang, D. Cheng, L. Zhang, ``Exploiting Motion Information from Unlabeled Videos for Static Image Action Recognition, '' in \textit{Proceedings of the AAAI conference on artificial intelligence}, 2020, pp. 12918-12925

\bibitem{ref28}
S. Sun, Z. Kuang, L. Sheng, W. Ouyang, W. Zhang, ``Optical Flow Guided Feature: A Fast and Robust Motion Representation for Video Action Recognition,'' in \textit{Proceedings of the IEEE conference on computer vision and pattern recognition}, 2018, pp. 1390-1399

\end{thebibliography}
\end{document}